\documentclass[conference]{IEEEtran}
\IEEEoverridecommandlockouts
\usepackage{cite}
\usepackage{amsmath,amssymb,amsfonts}
\usepackage{algorithmic}
\usepackage{graphicx}
\usepackage{textcomp}
\usepackage{xcolor}
\usepackage{hyperref}
\def\BibTeX{{\rm B\kern-.05em{\sc i\kern-.025em b}\kern-.08em
    T\kern-.1667em\lower.7ex\hbox{E}\kern-.125emX}}
\begin{document}

\title{TActiLE: Tiny Active LEarning for wearable devices\\
\thanks{*Equal contribution}
}

\author{\IEEEauthorblockN{1\textsuperscript{st} Massimo Pavan*}
\IEEEauthorblockA{\textit{DEIB} \\
\textit{Politecnico di Milano}\\
Milano, Italy \\
massimo.pavan@polimi.it}
\and
\IEEEauthorblockN{2\textsuperscript{nd} Claudio Galimberti*}
\IEEEauthorblockA{\textit{DEIB} \\
\textit{Politecnico di Milano}\\
Milano, Italy \\
claudio1.galimberti@mail.polimi.it}
\and
\IEEEauthorblockN{3\textsuperscript{rd} Manuel Roveri}
\IEEEauthorblockA{\textit{DEIB} \\
\textit{Politecnico di Milano}\\
Milano, Italy \\
manuel.roveri@polimi.it}
}

\maketitle

\begin{abstract}

Tiny Machine Learning (TinyML) algorithms have seen extensive use in recent years, enabling wearable devices to be not only connected but also genuinely intelligent by running machine learning (ML) computations directly on-device. Among such devices, smart glasses have particularly benefited from TinyML advancements. TinyML facilitates the on-device execution of the inference phase of ML algorithms on embedded and wearable devices, and more recently, it has expanded into On-device Learning (ODL), which allows both inference and learning phases to occur directly on the device. The application of ODL techniques to wearable devices is particularly compelling, as it enables the development of more personalized models that adapt based on the user’s data. However, one of the major challenges of ODL algorithms is the scarcity of labeled data collected on-device. In smart wearable contexts, requiring users to manually label large amounts of data is often impractical and could lead to user disengagement with the technology.

To address this issue, this paper explores the application of Active Learning (AL) techniques,  i.e., techniques that aim at minimizing the labeling effort, by actively selecting from a large quantity of unlabeled data only a small subset to be labeled and added to the training set of the algorithm. In particular, we propose TActiLE, a novel AL algorithm that selects from the stream of on-device sensor data the ones that would help the ML algorithm improve the most once coupled with labels provided by the user. TActiLE is the first Active Learning technique specifically designed for the TinyML context. We evaluate its effectiveness and efficiency through experiments on multiple image classification datasets. The results demonstrate that, within the stringent resource constraints of TinyML and wearables environments, TActiLE outperforms both a simple random selection baseline and more complex Stream Active Learning algorithms, demonstrating its suitability for tiny and wearable devices.
\end{abstract}


\section{Introduction}
\label{sec: intro}

Tiny Machine Learning (TinyML) is a fast-growing field related to artificial intelligence, focusing on applications and technologies for extremely low-profile devices. It allows to push intelligence to the edge reducing latency, power consumption, and bandwidth usage. TinyML enables efficient inference on low-power devices \cite{7}, but its capabilities can be further enhanced through On-Device Learning, which allows models to learn and evolve locally. Combining the two paradigms facilitates higher responsiveness, privacy-preserving, and adaptability of the deployed models.

TinyML algorithms have been successfully employed in many smart wearable devices, including also smart eyewear\cite{10.1007/978-3-031-71518-1_24}. In this context, On-device Learning techniques are deemed extremely interesting\cite{shalby2025dendronenhancinghumanactivity} because they enable the personalization of the models to the data of the final user of the wearable device and consent to continuously update of the model with the most up-to-date data collected directly from the user. In the eyewear context nevertheless, while there is the possibility of acquiring a lot of incoming data from the on-glasses sensors, the limited memory resources do not allow for storing a large amount of data. At the same time, continuously asking for a large amount of feedback from the user with the objective of obtaining labels for the models can be considered a bothersome task that must be reduced to a minimum during the operational life of the device. This task is made even more difficult by the lack of comfortable human-computer interfaces that characterize wearable devices in general and smart eyewear in particular. However, the presence of high-quality annotated data is one of the factors that make the success of deep learning models, and something that can not be ignored when designing On-device Learning algorithms for smart eyewear. 

For these reasons, Active Learning can be considered an extremely interesting paradigm to apply in the TinyML context. Active learning \cite{2} is a paradigm comprehending techniques by which a learning algorithm can accurately choose from an unlabeled set, a subset of the data to be labeled and used for training. The scope of these strategies is to reduce data annotation by selecting a subset of all the available samples for the model training, maintaining comparable performances with respect to the case in which the model is trained with the whole set of data. Classical active learning strategies struggle to handle high-dimensional data \cite{1} and since we focus on a TinyML context, we employed techniques of deep-active learning which are the results of the combination of deep learning and active learning strategies. The main active learning approach suited for this scenario is streaming-based sampling in which the data is scanned sequentially and the decision of taking the sample or discarding is made individually in an online manner, without the possibility of seeing again samples already discarded \cite{2}. 
Traditional active learning procedures, usually involve a one-by-one sampling procedure which leads to frequent retrainings, which is inefficient and can lead to overfitting. Morevover, since we are considering a deep learning architecture executing in a resource-constrained environment, we aim to reduce the number of retraining as much as possible. For these reasons, the deep-active learning strategy has to use a batch-based sampling strategy in which a set of samples is selected to be labeled before model retraining.
In this paper, we study how known and standard techniques of deep-active learning perform in a TinyML environment and we introduce \textit{TActiLE}, a novel active learning technique specifically designed for the TinyML context, that can work with two novel batch handling strategies.\\
The paper is organized as follows: in Chapter II we formulate the problem of deep stream active learning, in Chapter III we describe already known important aspects of deep active learning, in Chapter IV we formalize our solutions to address the problem, in Chapter V we show the comparison algorithms for our solutions and the parameters set used, in Chapter VI we show and comment empirical results, in Chapter VII we present the conclusions of the study and explores potential directions for future research.

\section{Problem formulation}
Let us consider an unlimited stream $S = \{(x_0, y_0), (x_1, y_1), \dots\} \subset X\times Y$ of data elements with relative labels. The initial subset $D_0 = \{(x_0, y_0),\dots,(x_T,y_T)\} \subset X_T \times Y_T$, with $T\in \mathbb{N}$ is used as a pre-training set for a deep neural network $f_0: X \to Y$. $f_0$ is the initial model used in a deep stream active learning procedure. The set of samples $S_U = \{x_{T+1}, x_{T+2}, \dots\}$ following $T$, assumed to be collected on-device, is missing the labels which can be queried to an omniscient oracle. The stream $S_U$ is the one subjected to the deep active learning procedure which, starting from the use of $f_0$, aims to satisfy the following conditions. At the generic $t$, given the current model $f_n$ trained on the dataset $D_n$, we want to decide whether to add or not $x_t$ to the to-be-labeled set $K_n$, in such a way that, at the end of the procedure, the set $K_n$ contains the $x^t_s$ that maximizes, with respect to all the other possible sets that could have been selected, the accuracy of the model $f_{n+1}$ trained on the dataset $D_{n+1} = D_n \cup K_n$. 
\newline
We want this selection to be performed in an online fashion, that is by deciding upon each sample reception if adding it to the batch $K_n$ or not. The procedure can only investigate the samples already inserted into the batch for the next query iteration, the ones discarded cannot be further evaluated.

\begin{figure}[t]
    \centering
    \includegraphics[width=0.45\textwidth]{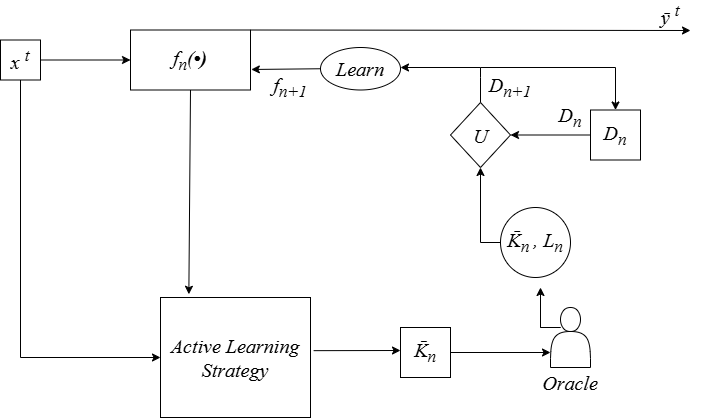}
    \caption{Visual representation of the problem formulation.}
    \label{fig:TactiLe_problemFormulation}
\end{figure}

\section{Related works}
\label{sec: related works}

\subsection{TinyML and On-device Learning}

TinyML solutions have demonstrated their effectiveness and efficiency in executing the inference phase of Machine and Deep Learning (MDL) models on resource-constrained devices \cite{warden_tinyml_2020, pavan2022tinyml}. However, the training of TinyML models is still largely assumed to take place in the Cloud, where ample computational and memory resources are available. This reliance on centralized training prevents TinyML models from continuously adapting to new data emerging in their operating environments.

On-device Learning (ODL) aims to bridge this gap by enabling both inference and learning phases to occur directly on edge devices. This advancement would allow intelligent TinyML-powered devices to dynamically adapt to nonstationary environments, shifting operational conditions, and user-specific preferences.

Historically, research in On-device Learning has focused on optimizing the training process for specific models, particularly neural networks \cite{lin_-device_2022, rueb_tinyprop_nodate, cai_tiny_2020}. Most of these studies assume a batch learning paradigm, where all training data are available in advance, allowing for multiple training epochs throughout the learning process.

More recent works, while still operating within batch learning constraints, have explored strategies for mitigating data storage limitations. Some have leveraged the "few-shot learning" paradigm \cite{10.1145/3703412.3703415, rusci_-device_2023}, developing algorithms capable of learning from minimal labeled data. Others have adopted a Stream Learning approach, processing incoming data in real-time to circumvent storage limitations. TinyML-based Stream Learning methods can be categorized into chunk-based approaches \cite{pavan2024tybox, ravaglia_tinyml_2021}, which use buffering techniques to retain compressed data representations, and fully online approaches \cite{ren_tinyol_2021, disabato_incremental_2020}, where each new data point is immediately utilized for learning and then discarded. An important aspect to consider when performing On-device Learning is the difficulties in performing any form of validation of the learned model\cite{falcetta2023personalize}.

Despite these advances, no existing work, to our knowledge, has explored Active Learning within the context of TinyML On-device Learning.

\subsection{Active Learning}
An active learning strategy starts with defining the query rules used to select the best set of samples from the unlabeled dataset. The main query strategies \cite{2} include uncertainty-based approaches which aim to select samples based on how much the model is uncertain, the greater the uncertainty the easier it is to be selected. Diversity-based strategies have the objective of maximizing the diversity between the selected samples. Hybrid query strategies can be defined as the combination of the two.

\section{Background on Stream Active Learning}
In a streaming deep-active learning procedure, two main central issues define the procedure's functioning: the trigger heuristic and the batch handling definition $m(\cdot)$. In particular, $m(\cdot)$ defines how each incoming sample $x_t$ is added to the set $K_n$ to be labeled, and the trigger heuristic decides when to use $K_n$ for the re-training of $f(\cdot).$ $\overline{K_n}$, the selected batch for the re-training, is sent to the oracle for labeling and concatenated to the already existing dataset to create $D_{n+1} = D_n \cup (\overline{K_n}, L_n)$. The resulting dataset is used to train $f_{n+1}$.

\subsubsection{Trigger heuristic}
Among the set of existing trigger heuristics, we have analyzed in depth a couple of the least requiring ones: the \textit{Windows heuristic} and the \textit{Batch Filling heuiristic}
\paragraph{Windows heuristic}
The first common way to express the trigger heuristic is by defining a fixed number of samples observed by the active procedure \cite{4}. The heuristic count the number of samples $d$ received after the retraining and triggers when $d = w$ where $w$ is the window length. Then the picked set of samples $\overline{K_n}$ is sent to an omniscient oracle for labelling. The batch $\overline{K_n}$ and the relative set of labels are added to the set collected so far $D_{n+1} = D_n \cup \overline{K_n}$. The model is finally retrained on the set $D_{n+1}$.

\paragraph{Batch Filling heuristic}
Another well-known trigger heuristic considers all the incoming samples received after the re-training without limiting the selection to a fixed number of data seen $w$, the part of the stream containing the samples arriving after the re-training is considered unlimited. This paradigm change reduces unnecessary sample selection since the active procedure is not restricted to a limited set. The heuristic triggers when the batch reaches the required length, $|K_n| = k$. The set $\overline{K_n}$ is labeled by the oracle and with the relative set of labels added to the set collected so far $D_{n+1} = D_n \cup \overline{K_n}$. The model is finally retrained on the set $D_{n+1}$.

\subsubsection{Batch handling}
To properly construct the batch $K$, each incoming sample $x_t$ is processed by batch handling functions, which, in turn, rely on metrics to assess the impact of adding the sample to the already existing batch. The batch handling functions return the best-performing batch according to those metrics which are expressed as objective functions. In literature, there is a well-known process called \textit{Submodular Maximization}, which aims to maximize the results of an objective function $m(K^t_n)$ in a stream context.
Submodular maximization algorithms make use of the results of an objective function as a metric to decide whether to add a sample to the batch or not. A known \textit{Submodular Maximization} that can be implemented for our deep-active learning purpose in tinyML context is \textit{preemption-streaming} \cite{6}. \textit{Preemption-streaming} implements $m(\cdot)$ in the following way. At each $t$, the batch handling technique outputs $K^{(t+1)}_n$, the batch at the next time instant. If $|K^t_n| < k$, $x^t$ is simply added to $K^t_n$, i.e., $ K^{(t+1)}_n = K^t_n \cup {x^t}$. If $|K^t_n| >= k$, the algorithm attempts to replace the incoming element with each element in the current batch. Every newly formed batch from these substitutions is evaluated using the objective function. If the best-performing batch achieves a higher value than the initial batch, it replaces the original, otherwise, the initial batch remains unchanged. Formally, we find $K^* = \arg\max_{K_i \in \{K_1, \dots, K_k\}} m(K_i)$, and, if the conditions $m(K^*) > m(K^t_n)$ holds, $K^{(t+1)}_n = K^*$. If the condition is not met, $K^{t+1}_n = K^t_n$.
The objective functions can differ depending on the query strategy used. Some of these metrics assess a sample's contribution based on the informativeness it provides to the model, computed using measures of the model's confidence in the sample, such as entropy \cite{4} or margin of confidence, evaluated based on the softmax probabilities. Other metrics assess the diversity of the incoming sample relative to the samples already added to the batch. The diversity is computed by evaluating some distance functions over a vectorial representation of the samples. There are also hybrid query strategies that combine the contribution of both the aforementioned metrics. This approach is very suited when considering a deep batch active learning algorithm since considering only an uncertainty-query strategy could lead to a batch composed of similar samples, providing redundant information, and decreasing its total meaningfulness. 

\section{TActiLE}
\textit{TActiLE} is our proposed solution to deal with deep active stream learning in a tiny context. TActiLE works with two novel batch-handling strategies $m(\cdot)$, \textit{Informativeness reference value} and \textit{Dual-mode reference value}.

\begin{figure}[t]
    \centering
    \includegraphics[width=0.45\textwidth]{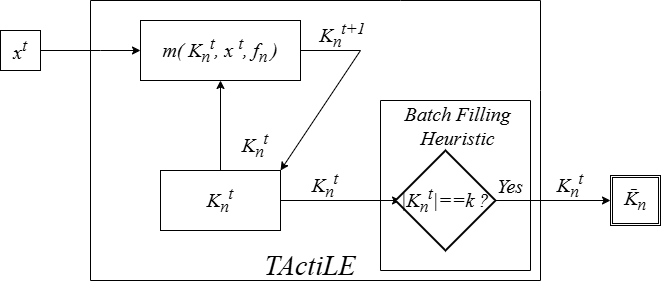}
    \caption{Visual representation of TActiLe batch handling procedure.}
    \label{fig:TactiLe_batchHandling}
\end{figure}

The first one implements similarly to other solutions \cite{2,1}, an uncertainty-based query strategy exploiting informativeness. The second one implements a hybrid query strategy by sequentially evaluating metrics of informativeness and diversity. As a heuristic trigger, TActiLE uses the aforementioned \textit{Batch Reaching} heuristic.

\subsection{Info RV}

\begin{figure}[h]
    \centering
    \includegraphics[width=0.45 \textwidth]{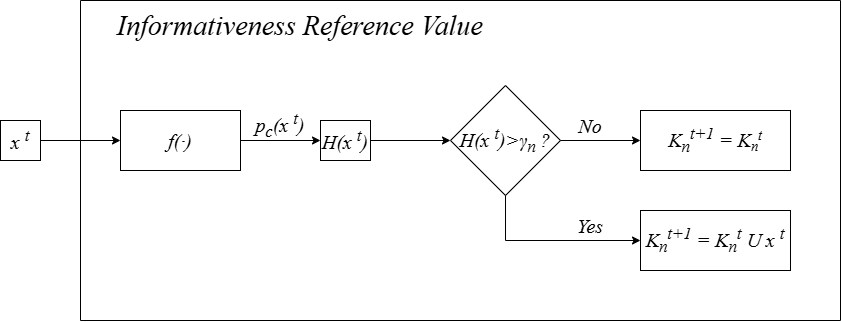}
    \caption{Visual representation of the informativeness reference value strategy.}
    \label{fig:infRV_zoom}
\end{figure}

\textit{Informativeness Reference Value} (Info RV) batch handling is performed by comparing the informativeness carried by the current sample $I(x^t)$ with a dynamic threshold $\gamma_n$. The metric used to assess the informativeness of samples is the entropy of the softmax probabilities predicted for the incoming sample. The metric evaluates the significant contribution that each sample individually possesses. At each $t$ the Info RV batch handling technique outputs $K^{(t+1)}_n$, the batch at the next time instant. Formally, if $I(x^t) > \gamma_n$ holds, $K^{(t+1)}_n = K^t_n \cup x^t$ otherwise $K^{(t+1)}_n = K^t_n$. Since \textit{Info RV} uses batch filling as a trigger heuristic, the procedure stops when $|K^t_n| = k$.
To have a reliable threshold on the most recent model, $\gamma_n$ is pre-computed before the beginning of the procedure and after every model retraining. It is calculated by evaluating $I(x^t)$ on the first $l$ datum received after each re-training. $\gamma_n$ is computed as the average of the $I(x^t)$ produced by the $j < l$ higher values. $l$ and $j$ are two hyperparameters that can be tuned depending on the data complexity that has to be processed. Furthermore, these two parameters control the selection frequency, in particular, the higher the $l$ the higher the precision of informativeness threshold $\gamma_n$ because it is evaluated over a higher number of samples. The $j$ parameter expresses the percentage of the calculated samples' informativeness effectively used to produce the threshold, the lower its value the stricter the condition expressed by the threshold.

\subsection{Dual RV}

\begin{figure}[h]
    \centering
    \includegraphics[width=0.45\textwidth]{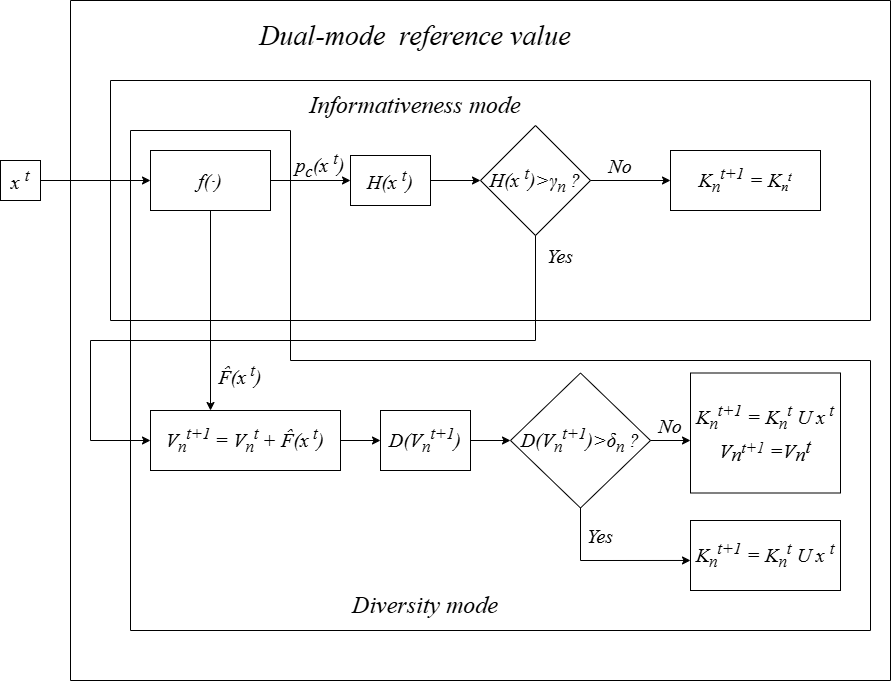}
    \caption{Visual representation of the Dual-mode reference value strategy.}
    \label{fig:dualMode_zoom}
\end{figure}

\textit{Dual-mode reference value} (Dual RV) is a batch handling procedure that uses as a metric a combination of uncertainty-based and diversity-based query strategies. The batch handling is performed by sequentially comparing the meaningfulness of the samples in terms of informativeness and diversity with relative thresholds $\gamma_n$ and $\delta_n$. The metric used for the informativeness evaluation is the same as the one used in \textit{Info RV}. The diversity metric instead produces a diversity value using $V^t_n$, which is composed of the vectorial representations of each data stored in $K^t_n$. In particular whenever a datum $x^t$ is added to $K^t_n$, its vectorial representation $\hat{F}(x^t)$, where $\hat{F}(\cdot)$ is the convolutional feature extractor of $f(\cdot)$, gets added to $V^t_n$. The diversity value $D(V^t_n)$ is computed as the sum of cosine distances between the vectorial representation of each sample couple, normalized by the number of possible pairs. \textit{Dual RV} starts comparing the informativeness carried out by the incoming sample with the relative threshold. If the condition is met the vectorial representation of the sample is added to the vectorial representations of the samples already in the batch. The diversity measure provided by the newly constructed vectorial representations $V^t_n$ is compared with the relative threshold and if even this second condition is met, the sample is added to the batch, otherwise, it is discarded and its vectorial representation is removed from $V^t_n$. Formally, at each $t$ the Dual RV batch handling technique outputs $K^{(t+1)}_n$, the batch at the next time instant. If $I(x^t) > \gamma_n$ holds, $\hat{F}(x^t)$ is added to $V^t_n$ producing $V^{(t+1)}_n = V^t_n \cup \hat{F}(x^t)$. If $D(V^{(t+1)}_n) > \delta_n$, $K^{(t+1)}_n = K^t_n \cup x^t$, otherwise $K^{(t+1)}_n = K^t_n$ and $V^{t+1}_n = V^t_n$.
Similarly to the \textit{Info RV} implementation, the diversity threshold is computed before the beginning of the process and after every model retraining. $\delta_n$ is calculated producing the vectorial representation $V_l$ of the first $l$ datum received after each re-training. The algorithm then, for $r$ times, picks a subset $R_q \subset V_l$ of size $q$ computing its diversity value. $\delta_n$ is the mean of the $j$ best diversity results. $l$, $r$, $q$, and $j$ are hyperparameters required by Dual RV strategy.

\section{Experimental setting}

In this section, first, we introduce the comparisons used in the experimental section, then we proceed to describe the datasets, parameters, and evaluation metrics used in the experiments, and finally, we describe the ML algorithms that will be used in conjunction with the strategies to evaluate them.

\subsection{Comparisons}

\subsubsection{Preemption\cite{4}} 

the \textit{Baseline with preemption} (Preemption) uses as heuristic trigger the \textit{Windows approach} strategy, and as batch handling procedure, it uses the aforementioned \textit{Preemption-streaming} algorithm, which in turn uses as metric a hybrid query strategy defined as the liner combination of an uncertainty-based query strategy and a diversity-based one. The objective function is defined as $g_p(K') = \lambda_ig_i(K')+\lambda_dg_d(K')$ where $\lambda_i$ and $\lambda_d$ are hyperparameters for scaling, $g_i(K')  = \sum_{x \in K'} \sum_{i} - P(x)_i \log P(x)_i $ and $g_d(K') = \frac{1}{2}logdet(I + \alpha A_s)$ requiring $A_s = [a(v_i,v_j)]_{i,j}$ is the similarity matrix, in which $a(v_i,v_j) = \langle \hat{F}(v_i), \hat{F}(v_j)\rangle$ represents the similarity measure between the vectorial representation of the pair of samples $x_i, x_j$ in the batch $K'$.

\subsubsection{Random sampling}
The random sampling query strategy picks randomly $k$ samples by the incoming stream without using any active learning procedure. As trigger heuristics uses the \textit{batch reaching full capacity} strategy.

\subsection{Datasets and Parameters}
The experiments were conducted on an Intel Core i5-9300H CPU 2.40GHz with one core usage setting. We carried out the experiments on three datasets commonly used in TinyML for computer vision tasks: 

\begin{itemize}
    \item MNIST\cite{lecun1998gradient} The MNIST dataset consists of 70,000 grayscale images of handwritten digits (0-9), each sized 28×28 pixels. It is widely used for benchmarking image classification models in machine learning.
    \item fashion-MNIST\cite{xiao2017fashion} Fashion-MNIST contains 70,000 grayscale images of 10 clothing categories, each with a resolution of 28×28 pixels. It serves as a more challenging alternative to the traditional MNIST dataset.
    \item CIFAR10\cite{krizhevsky2009learning} The CIFAR-10 dataset comprises 60,000 color images of size 32×32 pixels, categorized into 10 distinct classes. It is commonly used for training and evaluating deep learning models in image classification tasks.
\end{itemize}

All the experiments have been repeated 10 times varying the sample displacement in the dataset. The graphs show the mean and variance of the obtained results. We used an initial dataset $|D_0| = 150$ for MNIST, $|D_0| = 200$ for fashion-MNIST and $|D_0| = 3000$ for CIFAR10. 

The $k$ parameter is expressed as the concatenation of multiple sub-batches \cite{4} when considering the \textit{Preemption} because its execution time grows exponentially when using a higher batch size, and when considering the CIFAR10 dataset, because the initial training set for training the model on this dataset contains significantly more data compared to initial models trained on other datasets, and adding only a few samples for each retraining leads to overfitting.

The parameters used in the \textit{Preemption}, \textit{TActiLE + Info RV} and \textit{Tactile + Dual RV} strategies are reported in what follows.

\subsubsection{Preemption}

\begin{itemize}
    \item MNIST: $w=256$, $k = 16\times 2$
    \item fashion-MNIST: $w=256$, $k = 16 \times2$
    \item CIFAR10: $w=96$, $k = 16 \times8$
\end{itemize}

\subsubsection{TActiLE + Info RV}

\begin{itemize}
    \item MNIST: $l = 100$, $j = 25$, $k = 32$
    \item fashion-MNIST: $l = 25$, $j = 32$, $k = 32$
    \item CIFAR10: $l=100$, $j = 25$, $k = 32 \times 4$
\end{itemize}

\subsubsection{TActiLE + Dual RV} 

Informativeness computation: 

\begin{itemize}
    \item MNIST: $l = 100$, $j = 25$, $k = 32$
    \item fashion-MNIST: $l = 100$, $j = 75$, $k = 32$
    \item CIFAR10: $l=100$, $j = 75$, $k = 32 \times 4$\\
\end{itemize}

Diversity computation:

\begin{itemize}
    \item MNIST: $l=50$, $j=30$, $q=10$, $r=30$, $k=32$
    \item fashion-MNIST: $l=100$, $j=50$, $q=10$, $r=100$, $k=32$
    \item CIFAR10: $l=100$, $j=5$, $q=10$, $r=100$, $k=32 \times 4$
\end{itemize}

\subsection{Evaluation of the strategies}

We evaluate the performance of each strategy with three different metrics: Accuracy, time of execution, and memory requirements.

We evaluate the Accuracy performance of each strategy after each re-training of the ML algorithm, by measuring the accuracy of the ML algorithm on the respective test set of each dataset. The time of execution of each algorithm was measured as the average time required for the strategy to decide whether to add an incoming sample to the to-be-labeled set or discard it. The memory requirements of each strategy were measured as the additional memory overhead required for the execution of the strategy. For all algorithms, for both memory and time requirements, we included only the requirements of the strategy, leaving the requirements of the Neural Network outside of the measurement.

\subsection{Neural network architectures}

\begin{table}[t]
    \centering 
    \begin{tabular}{|c|c|c|}
    \hline
        \textbf{\textit{Layer}} & \textbf{\textit{Principal parameters}} & \textbf{\textit{Output Shape}}  \\
        \hline\hline
        \text{Conv2D} & Filters: 8, Kernel: (3x3) & (26, 26, 8)  \\
        \hline
        \text{MaxPooling2D}& Pool size: (2x2) & (13, 13, 8)  \\
        \hline
        \text{Conv2D } & Filters: 8, Kernel: (3x3) & (11, 11, 8)  \\
        \hline
        \text{MaxPooling2D } & Pool size: (2x2) & (5, 5, 8) \\
        \hline
        \text{Flatten} & - & (200)  \\
        \hline
        \text{Dense } & Neurons: 8 & (8)  \\
        \hline
        \text{Dense } & Neurons: 10 (classes) & (10)  \\
        \hline
    \end{tabular}
    \\[10pt]
    \caption{CNN architecture used for mnist dataset.}
    \label{tab:mnist_cnn}
\end{table}

\begin{table}[t]
    \centering 
    \begin{tabular}{|c|c|c|}
    \hline
        \textbf{\textit{Layer}} & \textbf{\textit{Principal parameters}} & \textbf{\textit{Output Shape}} \\
        \hline\hline
        \text{Conv2D} & Filters: 16, Kernel: (3x3) & (26, 26, 16)  \\
        \hline
        \text{MaxPooling2D}& Pool size: (2x2) & (13, 13, 8)  \\
        \hline
        \text{Conv2D } & Filters: 32, Kernel: (3x3) & (11, 11, 32)  \\
        \hline
        \text{MaxPooling2D } & Pool size: (2x2) & (5, 5, 32)  \\
        \hline
        \text{Conv2D } & Filters: 64, Kernel: (3x3) & (3, 3, 64) \\
        \hline
        \text{Flatten} & - & (576)  \\
        \hline
        \text{Dense } & Neurons: 64 & (64) \\
        \hline
        \text{Dropout } & Percentage: 0.25 & (64) \\
        \hline
        \text{Dense } & Neurons: 10 (classes) & (10)\\
        \hline
    \end{tabular}
    \\[10pt]    
    \caption{CNN architecture used for fashion-mnist dataset}
    \label{tab:fashionMNIST_cnn}
\end{table}

\begin{table}[t]
    \centering 
    \begin{tabular}{|c|c|c|}
    \hline
        \textbf{\textit{Layer}} & \textbf{\textit{Principal parameters}} & \textbf{\textit{Output Shape}} \\
        \hline\hline
        \text{Conv2D} & Filters: 32, Kernel: (3x3) & (30, 30, 32)  \\
        \hline
        \text{Conv2D} & Filters: 32, Kernel: (3x3) & (28, 28, 32)  \\
        \hline
        \text{MaxPooling2D}& Pool size: (2x2) & (14, 14, 32) \\
        \hline
        \text{Dropout } & Percentage: 0.3 & (14, 14, 32) \\
        \hline
        \text{Conv2D } & Filters: 64, Kernel: (3x3) & (12, 12, 64) \\
        \hline
        \text{Conv2D } & Filters: 64, Kernel: (3x3) & (10, 10, 64) \\
        \hline
        \text{MaxPooling2D } & Pool size: (2x2) & (5, 5, 64) \\
        \hline
        \text{Dropout } & Percentage: 0.4 & (5, 5, 64) \\
        \hline
        \text{Flatten} & - & (1600) \\
        \hline
        \text{Dense } & Neurons: 64 & (64) \\
        \hline
        \text{Dropout } & Percentage: 0.5 & (64) \\
        \hline
        \text{Dense } & Neurons: 10 (classes) & (10) \\
        \hline
    \end{tabular}
    \\[10pt]
    \caption{CNN architecture used for cifar10 dataset}
    \label{tab:cifar_cnn}
\end{table}

For every experiment, a different neural network architecture was deployed to handle the varying data complexity. The choice of each architecture was tailored to the characteristics of each dataset, allowing for generalization over the different scenarios. Table~\ref{tab:mnist_cnn}, Table~\ref{tab:fashionMNIST_cnn} and Table~\ref{tab:cifar_cnn} show respectively the architectures use for MNIST, fashion-MNIST, and CIFAR10.

The same training hyperparameters have been used for all the experiments. In particular, The models are trained using the Adam optimizer, which adaptively adjusts the learning rate during the training, improving convergence. As loss function has been used categorical cross-entropy, which is suitable for our objective of multiclass classification. The training is performed over 10 epochs. The batch size is set to 10 when considering MNIST and fashion-MNIST datasets, while when considering CIFAR10 it is set to 32. Early stopping was not implemented. Generic settings were used for both the training and the architecture's hyperparameters since the primary objective of this work was on the active learning selection procedure rather than on the gains provided by the architecture. The focus was placed on optimizing the model performance gains derived by the active learning query strategies, the model hyperparameter tuning was not examined in detail.

\section{Experimental results}
In this section, we report the graphs containing the experimental results obtained by the four query strategies \textit{Preemption}, \textit{TActiLE + Info RV}, \textit{TActiLE + Dual RV} and \textit{Random sampling}.\\

\subsection{Accuracy}

\begin{figure}[t]
    \centering
    \includegraphics[width=0.45\textwidth]{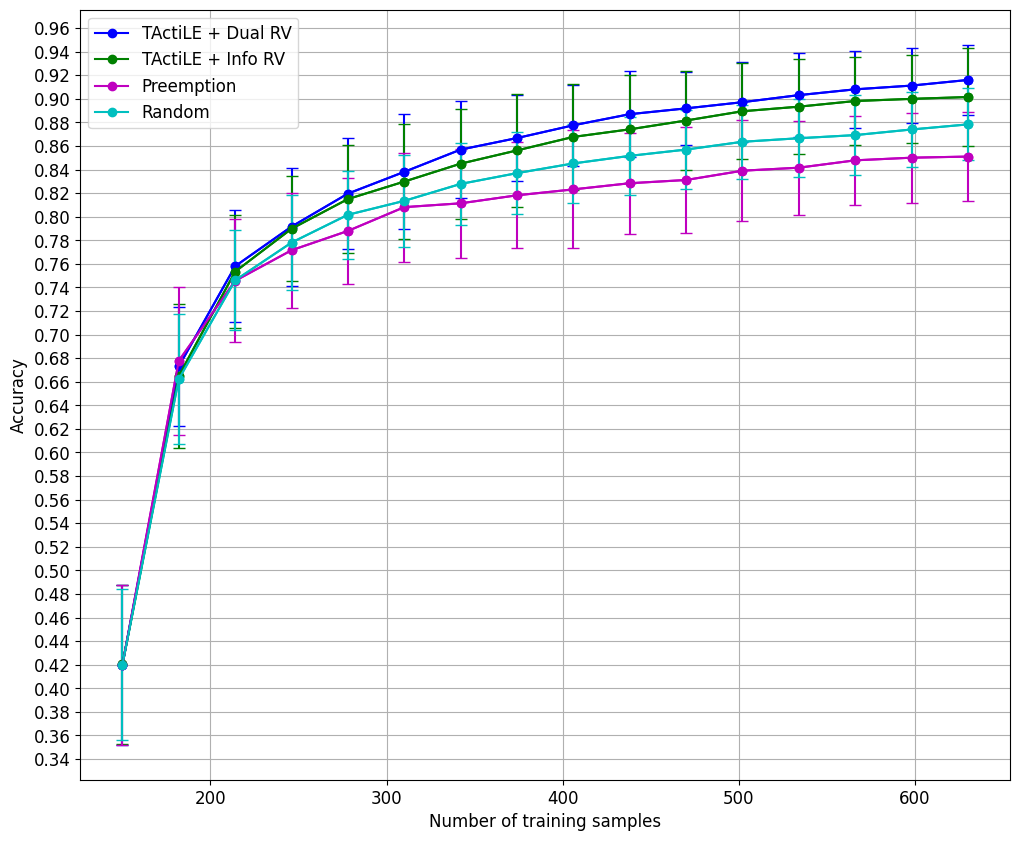}
    \caption{Model performances comparison on MNIST.}
    \label{fig:mnist}
\end{figure}

\begin{figure}[t]
    \centering
    \includegraphics[width=0.45\textwidth]{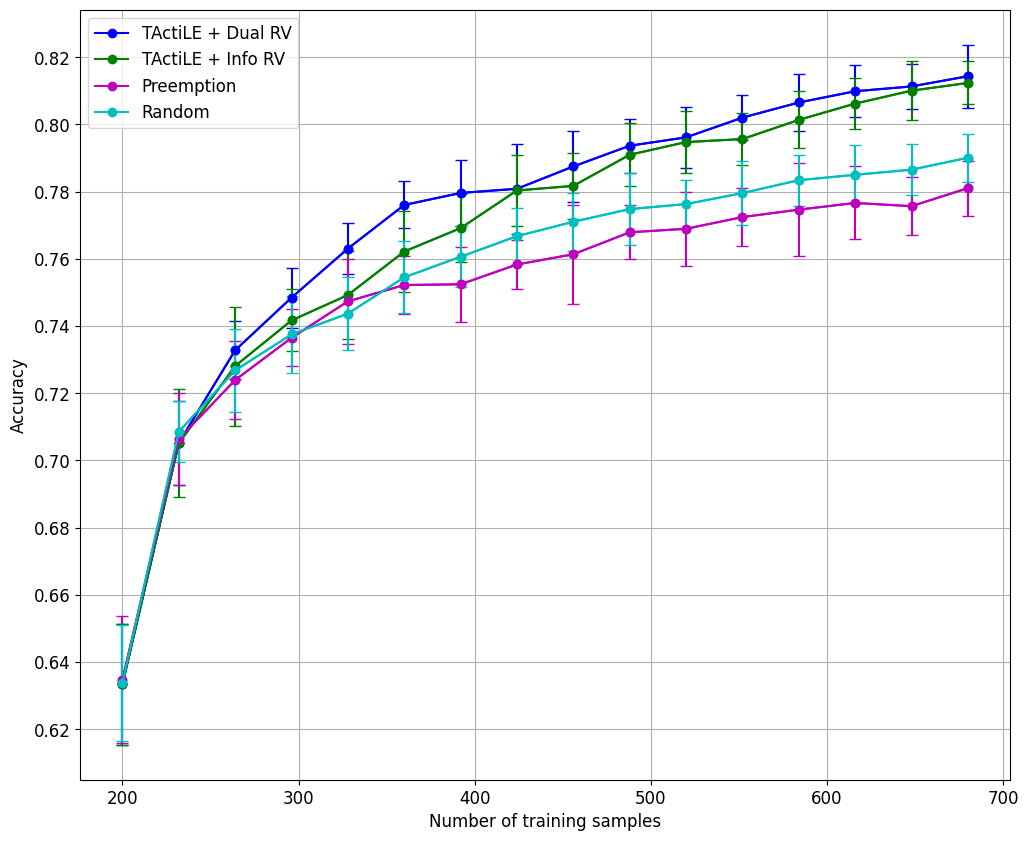}
    \caption{Model performances comparison on fashion-MNIST.}
    \label{fig:fashionMNIST}
\end{figure}

\begin{figure}[t]
    \centering
    \includegraphics[width=0.45\textwidth]{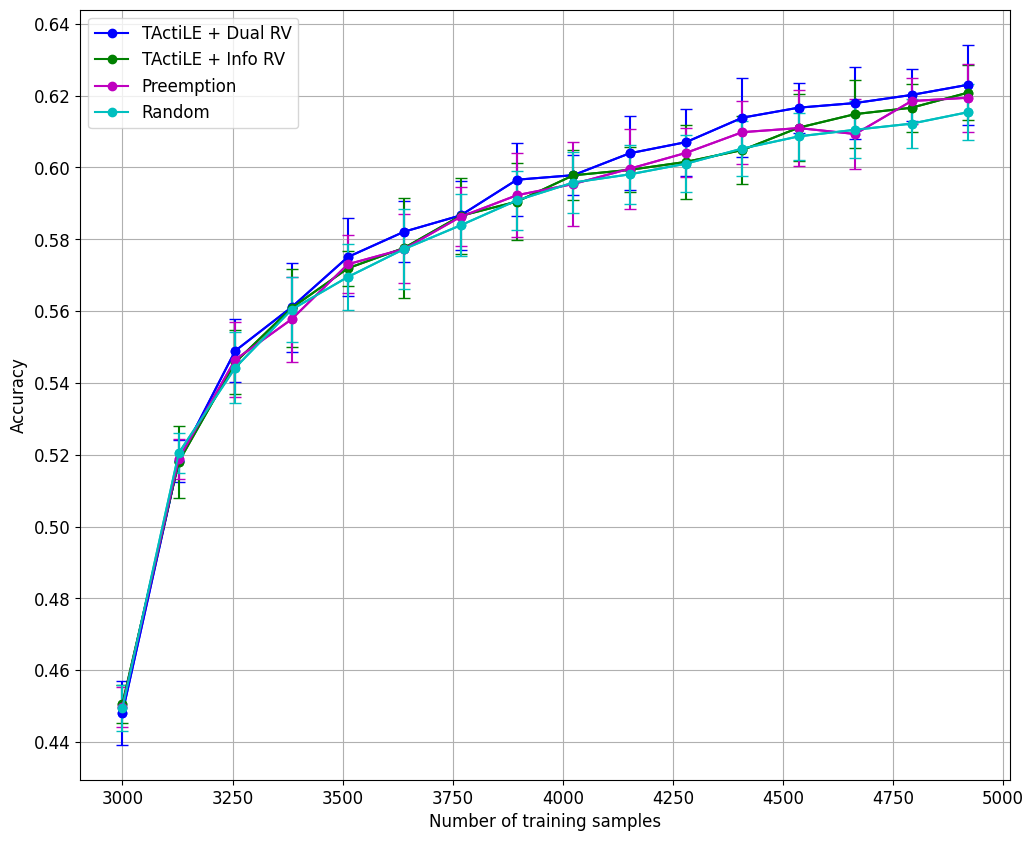}
    \caption{Model performances comparison on CIFAR10.}
    \label{fig:cifar}
\end{figure}

The accuracy results obtained by the query strategies on subsets of the MNIST, fashion-MNIST, and CIFAR10 datasets are reported in 
Figure~\ref{fig:mnist}, Figure~\ref{fig:fashionMNIST} and Figure~\ref{fig:cifar}, respectively. The figures report the accuracy obtained by the model after each retraining on $D_n$, obtained by labeling the data selected by the strategies.

The \textit{Preemption} algorithm has proven to be inaccurate in the TinyML context, where it needs to work with batches of extremely small size. Its accuracy results are lower than the one of the \textit{Random Baseline} considering MNIST and fashion-MNIST and slightly better considering CIFAR10.

Both the TActiLE-based solutions \textit{TActiLE + Dual RV} and \textit{TActiLe + Info RV} outperform both the \textit{Random baseline} and the \textit{Preemption} algorithm. Their main advantage lies in the fact that they allow the algorithm to evaluate many more samples before triggering the re-training, since they do not work on fixed-size windows. 
The two solutions obtain similar results on simple datasets, such as MNIST and fashion-MNIST, while the \textit{TActiLE + Dual RV} strategy obtains slightly better results with respect to \textit{TActiLE + Info RV} on more complex data and tasks, as happens in the CIFAR10 dataset.

\subsection{Time measurements and estimates on consumer MCU}

To further highlight the efficiency of the TActiLE approach over the comparisons, we have measured the time requirements of the strategies in the experimental setup on the fashion-mnist dataset, and we have estimated the time requirements on a consumer MCU, the Arduino Nicla Vision\cite{arduino_nicla_vision}. Results are reported in Table~\ref{table:velocities} and Table~\ref{table:velocitiesNicla}, for the measurements and estimates, respectively. The reported results are the average required by each algorithm to decide whether to add the new datum to the to-be-labeled set or discard it.

\begin{table}[h]
    \centering 
    \begin{tabular}{|c|c|c|c|}
    \hline
    \textbf{$k$} & \textbf{Preemption} & \textbf{TActiLE + Info RV} & \textbf{TActiLE + Dual RV} \\
    \hline \hline
    \textbf{16} & 2.5593 \textit{s} & 0.1504 \textit{s} & 0.2599 \textit{s} \\
    \textbf{32} & 10.5319 \textit{s}& 0.1494 \textit{s}& 0.2696 \textit{s}\\
    \hline
    \end{tabular}
    \\[10pt]
    \caption{Strategies' computational time, for analyzing one sample of fashion-mnist with different batch sizes.}
    \label{table:velocities}
\end{table}

In the tables, it is shown that the \textit{Preemption} algorithm is much slower than the other proposed solutions. Its computational time is directly related to the selected batch size and increasing the batch size exponentially increases the execution time of the algorithm. This is due to the fact that \textit{Preemption} at each iteration computes the determinant of the matrix of the cosine distances, and because, differently from TActiLE approaches, it almost always works with the full buffer. The results of the two TActiLE approaches, instead, show small increments in the execution times when augmenting the size of the batches. Furthermore, for both the tested batch sizes, the TActile + Info RV and TActiLE + Dual RV are at least one order of magnitude faster than the state-of-the-art approach preemption.

\begin{table}[h]
    \centering 
    \begin{tabular}{|c|c|c|c|}
    \hline
    \textbf{$k$} & \textbf{Preemption} & \textbf{Info RV} & \textbf{Dual RV} \\
    \hline \hline
    \textbf{16} & 12.7965 \textit{s} & 0.7520 \textit{s}& 1.2795 \textit{s} \\
    \textbf{32} & 110,1255 \textit{s}& 0.7470 \textit{s}& 1.3480 \textit{s} \\
    \hline
    \end{tabular}
    \\[10pt]
    \caption{Computational estimated time, for analyzing one sample of fashion-mnist with different batch sizes on the \textit{Nicla Vision}.}
    \label{table:velocitiesNicla}
\end{table}

The computational time required to run the algorithm on a \textit{Nicla Vision}, as shown in \textit{Table~\ref{table:velocitiesNicla}}, is estimated by dividing the clock rate of the device used for running the test by the clock rate of the \textit{Nicla Vision}. This is a rough estimate, and the results on the real device may slightly change. The clock rate of the PC used for the tests is 2.40 GHz, while the clock rate of the \textit{Nicla Vision} is 480 MHz, 0.48 GHz. The proportional factor to be used for the estimates is $\frac{2.40 \text{ GHz}}{0.48\text{GHz}}= 5$.

We estimate that the device could run the TActiLE + Info RV solution in under 1 second, while the TActiLE + Dual RV would require between 1 and 1.5 seconds to be executed on-device. Conversely, the Preemption approach would require at least 10 seconds to be executed on-device with a batch size of 16, and over 100 with a batch size of 32.

\subsection{Memory requirements}
In this paragraph are presented the memory requirements of all the analyzed algorithms. 
We will refer to the memory requirements of a single image as $M_I$, while $M_f$ will represent the memory requirements of a feature extracted by the convolutional feature extractor of the neural networks from the image. 

Since all the algorithms use the same dimension for the batch of samples they select, they also have similar requirements for the storage of this batch. The algorithms that compute the diversity among samples also require the storage of the latent features of the data in the batch, in order to compute the diversity faster. 

Since the \textit{Baseline with preemption} algorithm requires the constant swapping between a datum in the full buffer and the newly evaluated datum, it requires the storage (the batch dimension + the new datum under consideration) of $k+1$ images and $k+2$ (the batch dimension + the new datum under consideration and the copy of the feature being swapped out) latent features. In other words, the memory requirements $M_{preemption}$ of the \textit{preemption} algorithm can be computed as:

\[M_{preemption} = (k+1) \cdot M_I + (k+2) \cdot M_f \]

In the \textit{Dual-mode reference value} the diversity contribution of the incoming sample is calculated in attempting to insert its vectorial representation to the vectorial representation set of the already selected samples. This means that the space required to store the vectorial representation set is at most $k$ times the length of the vector produced by the vector extracted by the model's feature extractor. The memory requirements $M_{Dual}$ of the \textit{Dual-mode RV} algorithm can be computed as:

\[M_{Dual} = k \cdot M_I + k \cdot M_f\]


Since \textit{Informativeness reference value} and \textit{Random baseline} algorithms don't use a diversity metric as a query strategy, they don't require space to store any vectorial representation of that batch. In other words, the memory requirements $M_{info}$ and $M_{rand}$ can be computed as:

\[
M_{info} = k \cdot M_I \\
\]
\[
M_{rand} = k \cdot M_I
\]

For computing $M_I$ we have considered a precision of the image of 8 bits (1 byte) per pixel, the resulting $M_I$ for \textit{mnist} or \textit{fashion-mnist} is 784 bytes (28 x 28 x 1 x 1), and instead while it is 3072 bytes (32 x 32 x 3 x 1) for \textit{cifar10}. For the vectorial representations can be considered the use of floating point arrays, in which each element occupies 4 bytes. 

\begin{table}[h]
    \centering 
    \begin{tabular}{|c|c|c|c|}
    \hline
    \textbf{Dataset} & \textbf{Preemption} & \textbf{Dual RV} & \textbf{Info RV / Random} \\
    \hline \hline
    \textbf{\textit{mnist}} &53,072 KB &50,688 KB& 25,088 KB \\
    \textbf{\textit{fashion-mnist}} &104,208 KB&98,816 KB& 25,088 KB \\
    \textbf{\textit{cifar10}} &1.228,80 KB&1.212,93 KB& 393,728 KB \\
    \hline
    \end{tabular}
    \\[10pt]
    \caption{Memory requirements of all the algorithms considering a batch dimension $k=32$ for mnist e fashion-mnist e $k=128$ per cifar10.}
    \label{table:memEst_values}
\end{table}

For \textit{mnist} and \textit{fashion-mnist}, as reported in experimental settings, it was considered a batch size $k=32$, while for \textit{cifar10} a batch size of $k = 128$ was considered. The memory requirements of each algorithm are reported in \textit{Table~\ref{table:memEst_values}}.

\section{Conclusions}

This study focused on enabling, for the first time in the literature, the execution of Active Learning strategies on tiny devices, contributing to the TinyML and the on-device learning fields. Considering the limited computational and memory resources, learning in tiny scenarios can benefit a lot from an accurate selection of the samples on which the model is trained. This research introduced TActiLE, a new batch-handling strategy to actively select incoming samples in a stream. The proposed solution can work with two different query strategies, \textit{Info RV}, which implements an uncertainty-only metric, and \textit{Dual RV}, which combines both uncertainty-based and diversity-based metrics. TActiLE demonstrated superior performance with respect to a Random Baseline and to a state-of-the-art technique.

In light of the experimental results obtained, the \textit{TActiLE + Info RV} solution proved to be the most suitable when considering very constrained devices for analyzing simple data. However, with more complex data, evidence shows that a diversity metric can improve global performances, so with sufficient computational power, \textit{TActiLE + Dual RV} can be considered a competitive solution. 

Further advancements in selection strategies can enhance the accuracy of models trained on queried samples, particularly by refining diversity-based metrics. Future work will focus on improving these metrics to further bridge the gap between hybrid query strategies and uncertainty-based approaches. Additionally, exploring various data types—such as acoustic, vibrational, and motion sensor data—will help assess the effectiveness of the proposed solutions across different domains. Finally, active learning techniques that incorporate non-deep-learning methods for sample selection will be developed to expand the range of applicable strategies.

\bibliographystyle{plain} 
\bibliography{bibliography.bib}

\begin{thebibliography}{10}

\bibitem{7}
Youssef Abadade, Anas Temouden, Hatim Bamoumen, Nabil Benamar, Yousra Chtouki, and Abdelhakim~Senhaji Hafid.
\newblock A comprehensive survey on tinyml.
\newblock {\em IEEE}, 2023.

\bibitem{arduino_nicla_vision}
Arduino.
\newblock Arduino nicla vision.
\newblock \url{https://store.arduino.cc/products/nicla-vision}.
\newblock Accessed: 2025-03-21.

\bibitem{6}
Niv Buchbinder, Moran Feldman, and Roy Shwartz.
\newblock Online submodular maximization with preemption.
\newblock {\em ACM Trans}, 2019.

\bibitem{cai_tiny_2020}
Han Cai, Chuang Gan, Ligeng Zhu, and Song Han.
\newblock Tiny {Transfer} {Learning}: {Towards} {Memory}-{Efficient} {On}-{Device} {Learning}.
\newblock {\em arXiv:2007.11622 [cs]}, July 2020.
\newblock arXiv: 2007.11622.

\bibitem{10.1007/978-3-031-71518-1_24}
Arianna De~Vecchi, Alice Scandelli, Federica Bossi, Benedetta~Caterina Casadei, Hazem Hesham~Yousef Shalby, Marco Boschi, and Federica Villa.
\newblock Efficient human activity recognition: Machine learning at the sensor level.
\newblock In Maurizio Valle, Paolo Gastaldo, and Ernesto Limiti, editors, {\em Proceedings of SIE 2024}, pages 213--220, Cham, 2025. Springer Nature Switzerland.

\bibitem{disabato_incremental_2020}
Simone Disabato and Manuel Roveri.
\newblock Incremental on-device tiny machine learning.
\newblock In {\em Proceedings of the 2nd International workshop on challenges in artificial intelligence and machine learning for internet of things}, pages 7--13, 2020.

\bibitem{falcetta2023personalize}
Alessandro Falcetta, Massimo Pavan, Stefano Canali, Viola Schiaffonati, and Manuel Roveri.
\newblock To personalize or not to personalize? soft personalization and the ethics of ml for health.
\newblock In {\em 2023 IEEE 10th International Conference on Data Science and Advanced Analytics (DSAA)}, pages 1--10. IEEE, 2023.

\bibitem{krizhevsky2009learning}
Alex Krizhevsky.
\newblock Learning multiple layers of features from tiny images.
\newblock Technical report, University of Toronto, 2009.

\bibitem{lecun1998gradient}
Yann LeCun, L{\'e}on Bottou, Yoshua Bengio, and Patrick Haffner.
\newblock Gradient-based learning applied to document recognition.
\newblock {\em Proceedings of the IEEE}, 86(11):2278--2324, 1998.

\bibitem{lin_-device_2022}
Ji~Lin, Ligeng Zhu, Wei-Ming Chen, Wei-Chen Wang, Chuang Gan, and Song Han.
\newblock On-{Device} {Training} {Under} {256KB} {Memory}, July 2022.
\newblock arXiv:2206.15472 [cs].

\bibitem{pavan2022tinyml}
Massimo Pavan, Armando Caltabiano, and Manuel Roveri.
\newblock Tinyml for uwb-radar based presence detection.
\newblock In {\em 2022 International Joint Conference on Neural Networks (IJCNN)}, pages 1--8. IEEE, 2022.

\bibitem{10.1145/3703412.3703415}
Massimo Pavan, Gioele Mombelli, Francesco Sinacori, and Manuel Roveri.
\newblock Tinysv: Speaker verification in tinyml with on-device learning.
\newblock In {\em Proceedings of the 4th International Conference on AI-ML Systems}, AIMLSystems '24, New York, NY, USA, 2025. Association for Computing Machinery.

\bibitem{pavan2024tybox}
Massimo Pavan, Eugeniu Ostrovan, Armando Caltabiano, and Manuel Roveri.
\newblock Tybox: An automatic design and code generation toolbox for tinyml incremental on-device learning.
\newblock {\em ACM Transactions on Embedded Computing Systems}, 23(3):1--27, 2024.

\bibitem{ravaglia_tinyml_2021}
Leonardo Ravaglia, Manuele Rusci, Davide Nadalini, Alessandro Capotondi, Francesco Conti, Luca Benini, and Luca Benini.
\newblock A {TinyML} {Platform} for {On}-{Device} {Continual} {Learning} with {Quantized} {Latent} {Replays}.
\newblock {\em IEEE Journal on Emerging and Selected Topics in Circuits and Systems}, pages 1--1, 2021.

\bibitem{ren_tinyol_2021}
Haoyu Ren, Darko Anicic, and Thomas Runkler.
\newblock {TinyOL}: {TinyML} with {Online}-{Learning} on {Microcontrollers}.
\newblock {\em arXiv:2103.08295 [cs, eess]}, April 2021.
\newblock arXiv: 2103.08295.

\bibitem{1}
Pengzhen Ren, Yun Xiao, Xiaojun Chang, Po-Yao Huang, Zhihui Li, Brij~B. Gupta, Xiaojiang Chen, and Xin Wang.
\newblock A survey of deep active learning.
\newblock {\em arXiv}, 2021.

\bibitem{rueb_tinyprop_nodate}
Marcus Rueb, Daniel Maier, Daniel Mueller-Gritschneder, and Axel Sikora.
\newblock {TinyProp} - {Adaptive} {Sparse} {Backpropagation} for {Efficient} {TinyML} {On}-device {Learning}.

\bibitem{rusci_-device_2023}
Manuele Rusci and Tinne Tuytelaars.
\newblock On-device {Customization} of {Tiny} {Deep} {Learning} {Models} for {Keyword} {Spotting} with {Few} {Examples}.
\newblock {\em IEEE Micro}, pages 1--7, 2023.
\newblock Conference Name: IEEE Micro.

\bibitem{4}
Yuya Senzaki and Christian Hamelain.
\newblock Active learning for deep neural networks on edge device.
\newblock {\em arXiv}, 2023.

\bibitem{2}
Burr Settles.
\newblock Active learning literature survey.
\newblock Technical report, University of Wisconsin--Madison, 2009.

\bibitem{shalby2025dendronenhancinghumanactivity}
Hazem Hesham~Yousef Shalby and Manuel Roveri.
\newblock Dendron: Enhancing human activity recognition with on-device tinyml learning, 2025.

\bibitem{warden_tinyml_2020}
Pete Warden and Daniel Situnayake.
\newblock {\em {TinyML}: Machine Learning with {TensorFlow} Lite on Arduino and Ultra-low-power Microcontrollers}.
\newblock O'Reilly, 2020.
\newblock Google-Books-{ID}: {sB}3mxQEACAAJ.

\bibitem{xiao2017fashion}
Han Xiao, Kashif Rasul, and Roland Vollgraf.
\newblock Fashion-mnist: a novel image dataset for benchmarking machine learning algorithms.
\newblock {\em arXiv preprint arXiv:1708.07747}, 2017.

\end{thebibliography}


\end{document}